\documentclass[runningheads]{llncs}
\usepackage[T1]{fontenc}
\usepackage[utf8]{inputenc}
\usepackage{graphicx}
\usepackage{hyperref}
\usepackage{color}
\usepackage{amsmath, amsfonts, amssymb}
\usepackage{marvosym}

\begin{document}

\title{MADTempo: An Interactive System for Multi-Event Temporal Video Retrieval with Query Augmentation} 
\titlerunning{MADTempo}
\pagestyle{empty}
%
\author{Huu-An Vu\inst{1} \and
Van-Khanh Mai\inst{2} \and
Trong-Tam Nguyen\inst{1} \and
Quang-Duc Dam\inst{1} \and
Tien-Huy Nguyen\inst{3} \and
Thanh-Huong Le\inst{1} 
\thanks{This research is supported by AI VIETNAM\cite{aivietnam2025}.}}
\begingroup
\renewcommand\thefootnote{\Letter}
\footnotetext{Corresponding author: huynt654@gmail.com.}
\endgroup
\authorrunning{Huu-An Vu et al.}
\institute{
Hanoi University of Science and Technology, Hanoi, Vietnam \\
\email{\{an.vh225467, tam.nt225527, duc.dq225483\}@sis.hust.edu.vn, huonglt@soict.hust.edu.vn}
\and
The University of Danang - University of Science and Technology, Danang, Vietnam \\
\email{102220194@sv1.dut.udn.vn}
\and
Vietnam National University, Ho Chi Minh City, Vietnam
}
\maketitle              

\begin{abstract}
The rapid expansion of video content across online platforms has intensified the need for retrieval systems capable of understanding not only isolated visual moments but also the temporal structure of complex events. Existing approaches often fall short in modeling temporal dependencies across multiple events and in handling queries that reference unseen or rare visual concepts. To address these challenges, we introduce \textbf{MADTempo}, a video retrieval framework developed by our team, \textbf{AIO\_Trình}, that unifies temporal search with web-scale visual grounding. Our temporal search mechanism captures event-level continuity by aggregating similarity scores across sequential video segments, enabling coherent retrieval of multi-event queries. Complementarily, a Google Image Search–based fallback module expands query representations with external web imagery, effectively bridging gaps in pretrained visual embeddings and improving robustness against out-of-distribution (OOD) queries. Together, these components advance the temporal reasoning and generalization capabilities of modern video retrieval systems, paving the way for more semantically aware and adaptive retrieval across large-scale video corpora.
\end{abstract}


\section{Introduction}
The rapid expansion of online video content has created massive multimedia archives, posing critical challenges in efficient storage, indexing, and retrieval. Vision-Language models (VLMs)~\cite{11147437,10.1145/3701716.3717653} have become potential direction to approach the problem. However, as thousands of hours of video are uploaded daily, users increasingly demand systems capable of locating specific moments or events through natural language queries. This has driven the development of advanced video retrieval pipelines that integrate image, audio language, and video understanding~\cite{nguyennhu2025stervlmspatiotemporalenhancedreference,nguyen2024improvinggeneralizationvisualreasoning}.

Modern systems~\cite{tran2025transforming,yamada2024video,rek2024nvidia,nguyennhu2025lightweightmomentretrievalglobal,tran2025efficientrobustmomentretrieval} typically employ large-scale vision-language models (VLMs) such as CLIP~\cite{radford2021clip}, BLIP-2~\cite{li2023blip2}, and VideoCLIP~\cite{xu2021videoclip} to align textual and visual semantics within a shared embedding space. To enhance fine-grained understanding, auxiliary modalities including Optical Character Recognition (OCR)~\cite{du2020ppocr,nguyen2019vietocr}, Object Detection (OD)~\cite{ren2015faster,liu2016ssd,redmon2016yolo,carion2020detr}, and Automatic Speech Recognition (ASR)~\cite{graves2006ctc,amodei2016deepspeech2,chan2016las,gulati2020conformer} are widely integrated to extract textual, object-level, and auditory cues. Temporal segmentation and cross-modal fusion techniques further enable modeling of event-level dynamics, forming the basis of state-of-the-art retrieval architectures.

Recent research and international challenges such as the Video Browser Showdown (VBS)~\cite{schoeffmann2013vbs}, Lifelog Search Challenge (LSC)~\cite{gurrin2024lsc}, and the Ho Chi Minh City AI Challenge 2025~\cite{AIChallenge2025} have driven progress in localized event retrieval. However, most existing methods remain limited to single-event or short-term localization~\cite{le2024fustar,long2024avisearch,nguyen2024artemissearch,tran2024interactive,vo2024vietnamese,vuong2024newsinsight}. Extending retrieval to multi-event queries—where multiple semantically related actions must be temporally ordered—introduces new challenges in maintaining semantic continuity and long-range temporal consistency.

In this paper, we introduce \textbf{MADTempo}, a video retrieval framework designed to address these limitations. This system served as the core submission for our team, \textbf{AIO\_Trình}, achieving a high-ranking result in the Ho Chi Minh City AI Challenge 2025. Our main contributions are threefold:
\begin{enumerate}
    \item We design an end-to-end video retrieval pipeline combining vision-language representations (e.g., CLIP-Laion~\cite{cherti2023openclip}) with efficient similarity search using Milvus~\cite{wang2021milvus} and MongoDB~\cite{mongodb2024}, augmented by a GPT-5-based~\cite{openai2025gpt5card} query enhancement module for richer semantic representations.
    \item We introduce a Google Image Search–based fallback mechanism to handle out-of-distribution (OOD) queries by augmenting unseen visual concepts with web-scale imagery.
    \item We propose a robust temporal search pipeline tailored for multi-event retrieval, ensuring scalable and coherent retrieval of complex event sequences across large video corpora.
\end{enumerate}

Together, these components enable fast, scalable, and temporally consistent retrieval, advancing the state of large-scale video understanding.

\section{Related work}
Recent advances in video retrieval have been driven by deep learning and multimodal architectures. Our work builds upon three central research pillars: (1) Vision-Language Models (VLMs) for cross-modal representation, (2) the integration of fine-grained multimodal information, and (3) temporal-based retrieval and localization methods.

Large-scale VLMs such as CLIP~\cite{radford2021clip}, BLIP-2~\cite{li2023blip2}, and VideoCLIP~\cite{xu2021videoclip} have transformed the field by learning joint embedding spaces for text and visual content, enabling direct matching between natural language queries and video segments. In our system, we adopt CLIP-Laion~\cite{cherti2023openclip} for enhanced generalization.

To achieve detailed semantic understanding, modern pipelines augment VLMs with auxiliary modalities. Automatic Speech Recognition (ASR) models such as DeepSpeech2~\cite{amodei2016deepspeech2} and Conformer~\cite{gulati2020conformer} extract spoken content, while Optical Character Recognition (OCR) systems such as PP-OCR~\cite{du2020ppocr} and VietOCR~\cite{nguyen2019vietocr} capture on-screen text. Object Detection (OD) models, including YOLO~\cite{redmon2016yolo} and DETR~\cite{carion2020detr}, identify salient entities. Integrating these modalities has become standard practice for achieving semantically rich video representations.

Temporal retrieval research extends these foundations toward identifying when events occur. Challenges such as the Video Browser Showdown (VBS), Lifelog Search Challenge (LSC), and the Ho Chi Minh City AI Challenge 2025~\cite{AIChallenge2025} have advanced methods for single-event localization~\cite{long2024avisearch,nguyen2024artemissearch,tran2024interactive,vo2024vietnamese,vuong2024newsinsight}. However, most existing systems degrade when handling multi-event tracking queries requiring long-range temporal consistency and semantic continuity.

At the system level, scalable retrieval increasingly relies on vector databases such as Milvus~\cite{wang2021milvus} for high-speed similarity search and NoSQL databases such as MongoDB~\cite{mongodb2024} for metadata management. Emerging research also explores Large Language Model (LLM)-based query refinement, where models such as GPT-5~\cite{openai2025gpt5card} generate diverse query variations to improve recall. 

Building on these insights, our work introduces a temporal search mechanism explicitly designed for multi-event retrieval and a Google Image Search–based fallback for robust handling of out-of-distribution (OOD) visual queries.

\section{System Baseline and Preprocessing Framework}

\subsection{Data Preprocessing}

\begin{figure}[h]
    \centering
    \includegraphics[width=\linewidth]{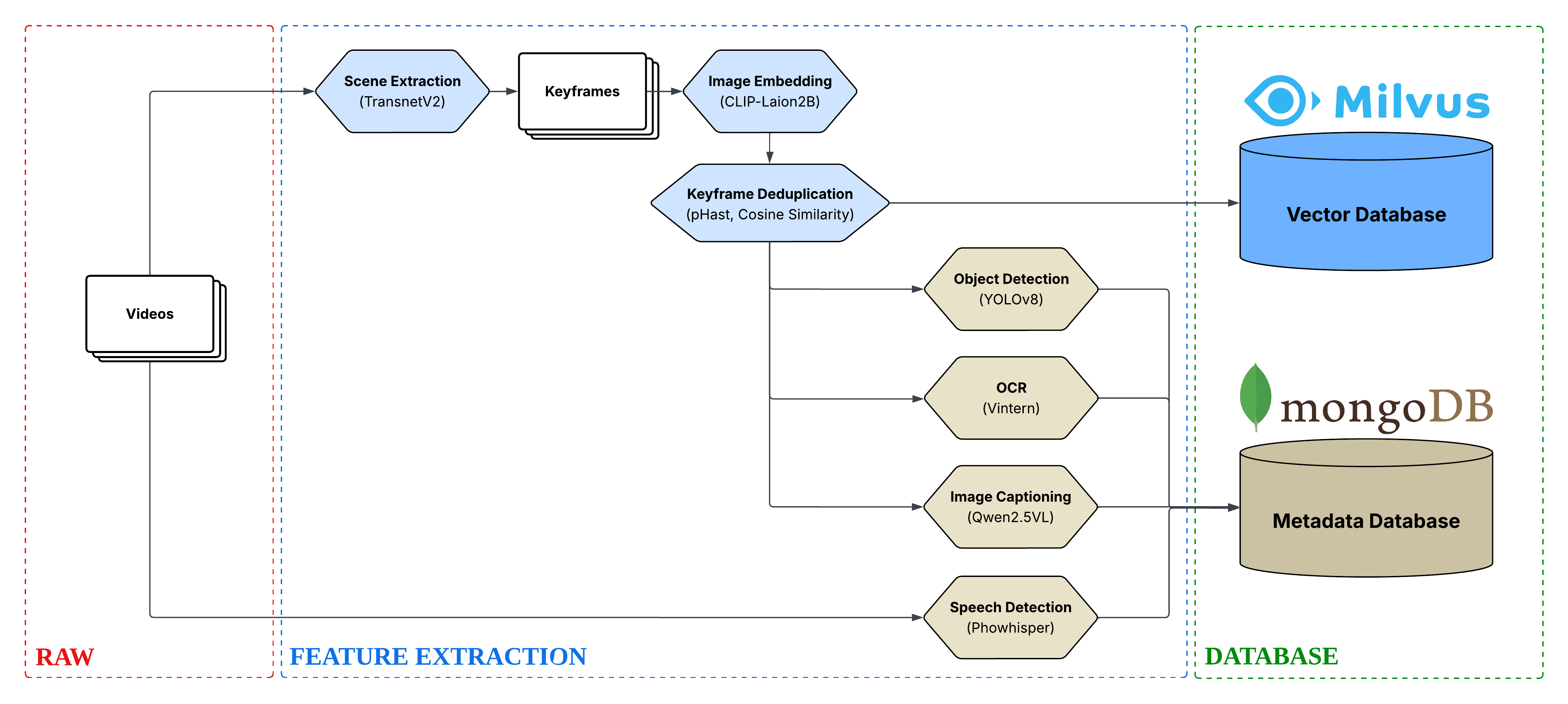}
    \caption{Overview of the data preprocessing and features extraction pipeline.}
    \label{fig:pipeline}
\end{figure}

The preprocessing pipeline (Figure~\ref{fig:pipeline}) converts raw videos into a structured multimodal database comprising two complementary components: a vector database for visual similarity search and a metadata database for text- and attribute-based retrieval.

\subsubsection{Keyframe Extraction}
Videos are segmented into shots using TransNetV2~\cite{soucek2020transnetv2}, from which representative keyframes are selected. Each keyframe is encoded using CLIP-Laion~\cite{cherti2023openclip} to obtain a semantic embedding $v \in \mathbb{R}^d$. A two-stage deduplication is applied to remove redundancy—first via perceptual hashing (pHash)~\cite{prathima2024phash}, then by cosine similarity filtering ($>0.965$) on embeddings. The remaining unique keyframes form the visual basis for indexing and analysis.

\subsubsection{Vector Storage}
All embeddings are indexed into Milvus~\cite{wang2021milvus}, enabling high-speed $k$-NN similarity search across millions of feature vectors.

\subsubsection{Multimodal Metadata Extraction}
To enrich visual understanding, each keyframe is analyzed across multiple modalities. YOLOv8~\cite{jocher2023yolov8} detects objects, Vintern-1b-v3\_5~\cite{vintern2024ocr} extracts on-screen text (OCR), and Qwen2.5VL~\cite{gwen2024caption} generates captions. Audio is transcribed via PhoWhisper~\cite{whisper2023}, a Vietnamese-adapted Whisper model for ASR. These outputs provide complementary semantic cues describing objects, scenes, and spoken content.

\subsubsection{Metadata Storage}
All extracted information—objects, text, captions, and transcripts—is stored in MongoDB~\cite{mongodb2024} under a unified schema. This repository supports structured filtering and full-text search, enabling fine-grained cross-modal retrieval and scalable semantic querying.

\subsection{System Overview}

\begin{figure}[h]
    \centering
    \includegraphics[width=\linewidth]{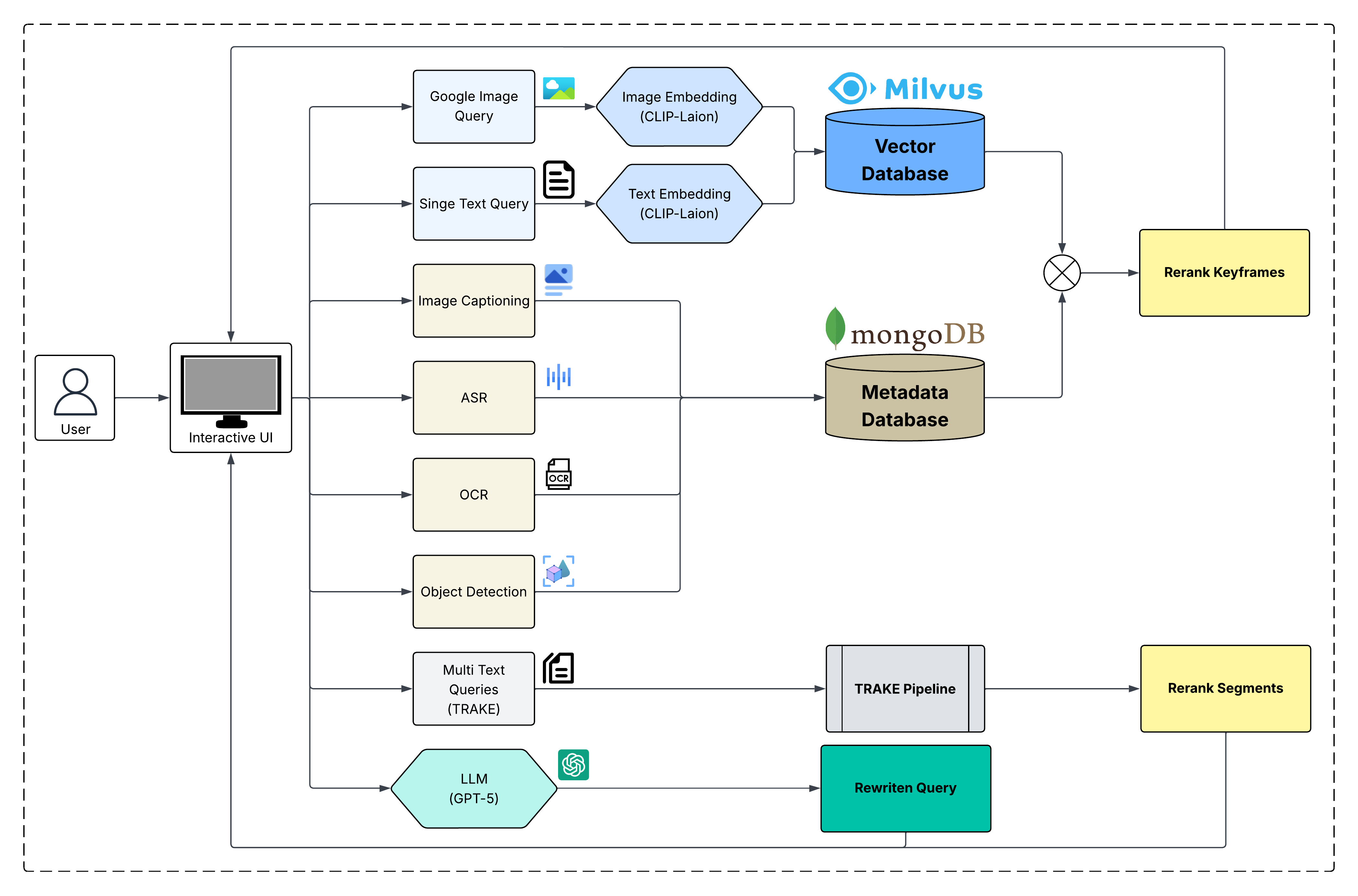}
    \caption{Overall system architecture illustrating the interaction between user interface, multimodal analysis modules, and dual-database retrieval framework. The system integrates CLIP-based vector search in Milvus and metadata filtering via MongoDB, coordinated through an interactive UI to support both simple and complex text queries.}
    \label{fig:system_architecture}
\end{figure}

Our system (Figure~\ref{fig:system_architecture}) supports multimodal and temporal video retrieval through a dual-database backend constructed during preprocessing. The interactive interface orchestrates two retrieval pipelines: keyframe-level retrieval and multi-event temporal retrieval (TRAKE).

\subsubsection{Keyframe-Level Retrieval}

This pipeline retrieves precise visual moments by combining vector similarity search with multimodal metadata filtering. A query vector $v_q$ is produced via one of two mechanisms.

\paragraph{Direct Text Query.}
A textual query $Q_t$ (e.g., ``a man wearing a red hat'') is encoded with the CLIP-LAION text encoder~\cite{radford2021clip}:
\[
v_q = \text{CLIP}_{\text{text}}(Q_t)
\]
Milvus~\cite{wang2021milvus} then performs similarity search over keyframe embeddings.

\paragraph{Context-Expanded Image Query.}
To handle rare or ambiguous concepts, Google Image Search retrieves top-$k$ images for $Q_t$. The user selects one representative image $I_{\text{selected}}$, which is encoded with the CLIP-LAION image encoder:
\[
v_q = \text{CLIP}_{\text{image}}(I_{\text{selected}})
\]

\paragraph{Hybrid Search and Filtering.}
The resulting $v_q$ is used for $k$-NN retrieval in Milvus (cosine similarity). Metadata filters in MongoDB refine results across ASR, OCR, and object-detection modalities. A join aligns Milvus candidates $\mathcal{K}$ with metadata constraints $Q_m$:
\[
\text{FinalSet} = \{k_j \in \mathcal{K} \mid \text{Filter}(M(k_j), Q_m)\}
\]
The final set is re-ranked by combined similarity and metadata relevance.

\subsubsection{Complex and Temporal Retrieval}

This pipeline retrieves dynamic events and multi-event structures. Users may specify explicit subevents or issue a single complex natural-language query $Q_{\text{raw}}$. In the latter case, GPT-5~\cite{openai2025gpt5card} parses $Q_{\text{raw}}$ into structured components:
\[
Q_{\text{raw}} \rightarrow \{Context, Event_1, \dots, Event_n\}
\]
Each event is embedded via CLIP-LAION and retrieved with Milvus, while temporal coherence is validated using multimodal metadata from MongoDB.

For queries involving explicit temporal relations (e.g., ``A after B''), the TRAKE pipeline performs structured temporal reasoning to identify sequences consistent with the specified constraints. Details appear in Section~\ref{sec:trake}.

\section{Methodology}
\subsection{TRAKE}
\label{sec:trake}

\begin{figure}[h]
    \centering
    \includegraphics[width=\linewidth]{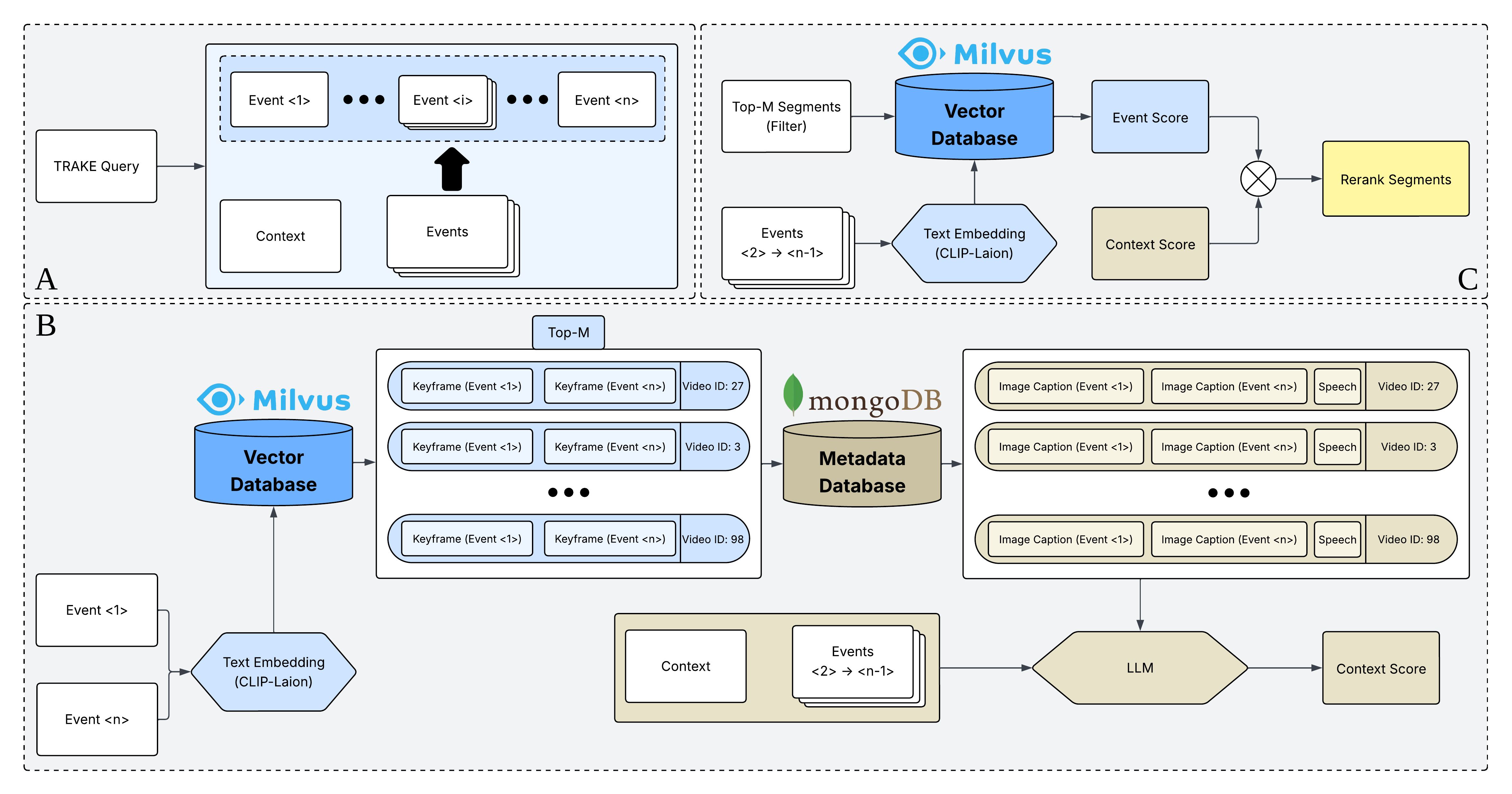}
    \caption{The TRAKE pipeline for multi-event temporal video retrieval. (A) Query decomposition via LLM or manual input. (B) Retrieval of candidate segments using boundary events $E_1$ and $E_n$, followed by context scoring via LLM. (C) Final reranking of segments using intermediate events $E_2 \to E_{n-1}$ and weighted fusion of event and context scores.}
    \label{fig:trake}
\end{figure}

We propose a TRAKE pipeline, demonstrated in Figure~\ref{fig:trake}, introducing a multi-event temporal search framework designed to retrieve video segments that jointly satisfy a sequence of event-based conditions described in a natural language query. Given a TRAKE query $Q$, the system decomposes it into two components: the context $C$ and an ordered sequence of events $E = \{E_1, E_2, ..., E_n\}$. The overall objective is to identify a video segment $S_i$ within a large-scale corpus that maximizes its semantic and temporal consistency with all specified events.

\begin{enumerate}
    \item \textbf{Query Decomposision:}
    The query $Q$ can either be automatically parsed using a large language model (LLM, e.g., GPT-5) or manually specified by the user. This stage produces a structured representation:
    \[
    Q \Longrightarrow \{C, E_1, E_2, \dots, E_n\}
    \]
    where $C$ captures the global context (e.g., “during a football match”), and each $E_i$ represents an event (e.g., “player kicks the ball,” “goalkeeper saves”). This structured representation serves as the basis for subsequent retrieval and ranking stages.
    \item \textbf{Context-Aware Score:}
    To localize candidate temporal segments, we compute similarity scores between each boundary event and the \textit{Keyframe Corpus} $\mathcal{K}$:
    \[
        SimScore_i(k) = sim(z_{E_i}, z_k), \quad where \quad z_x = f_{CLIP}(x)
    \]
    
    For every keyframe $k_1$ that matches event $E_1$, we search within the same video for feasible end keyframes that satisfy temporal ordering and a maximum duration constraint:
    \[
        K_n(k_1) = \{ k_n \} \quad  s.t.\quad
        \begin{cases}
            V(k_1)=V(k_n) \\
            0 < t_{k_n}-t_{k_1} \le (n-1)\cdot \tau
        \end{cases}
    \]
    
    where $\tau$ is the maximum allowed duration between consecutive events, and $t_k$ is the timestamp of the keyframe $k$ in video $V(k)$. This ensures temporal order and segment compactness. Among this set, we select the end keyframe with the highest similarity to $E_n$:
    \[
        \hat{k}_n = \arg\max_{k_n \in K_n(k_1)} SimScore_n(k_n).
    \]
    
    Each valid pair $(k_1, \hat{k}_n)$ defines a candidate segment:
    \[
        S_i = (k_1, \hat{k}_n, V_i).
    \]
    
    where $V_i$ is the video id of keyframes in segment $S_i$. We then compute a boundary confidence score for each segment:
    \[
        BoundaryScore(S_i) = SimScore_1(k_1) + SimScore_n(\hat{k}_n).
    \]
    
    All segments are ranked by $BoundaryScore$, and the top-$M$ highest-scoring candidates are retained for contextual evaluation. This procedure preserves all feasible starting points while selecting only the most semantically consistent end points, enabling efficient downstream analysis without discarding strong candidates.
    
    For each valid segment $S_i$, we extract multimodal metadata:
    \[
    Meta_i = (IC_1, IC_n, Speech_{1 \rightarrow n}, V_i)
    \]
    where $IC_1$ and $IC_n$ denote image captions for boundary keyframes, and $Speech_{1 \rightarrow n}$ represents ASR transcripts between them. The metadata is stored in MongoDB and then paired with the intermediate context and sub-events $\{E_1, E_2, \dots, E_n\}$. An LLM then evaluates the semantic coherence between the contextual sub-events and each candidate segment:
    \[
    ContextScore(S_i)=\frac{1}{100} \cdot f_{LLM}(Meta_i, C, \{E_1, E_2, \dots, E_n\})
    \]
    where $f_{LLM}(\cdot)$ outputs a score between 0–100, later normalized to [0, 1].
    \item \textbf{Event-Based Retrieval:}
    While the previous stage localizes coarse candidate segments based on boundary events and contextual coherence, this stage refines temporal alignment by enforcing the sequential occurrence of all intermediate events $\{E_2, \dots, E_{n-1}\}$ within each candidate segment.

    First, we construct a \textit{Filtered Keyframe Corpus} $\mathcal{K}_F$ that contains only keyframes belonging to the top-$M$ candidate segments obtained from the context-aware stage:
    \[
    \mathcal{K}_F = \{ k \mid k \in S_i, \, S_i \in \text{Top-}M \}.
    \]
    
    For each intermediate event $E_j$ ($2 \le j \le n-1$), we compute similarity scores between its embedding and every keyframe in the filtered corpus:
    \[
    SimScore_j(k) = sim(z_{E_j}, z_k), \quad \forall k \in \mathcal{K}_F.
    \]
    
    To ensure strict temporal ordering and intra-video consistency, we perform a beam search for each candidate segment $S_i = (k_1, \hat{k}_n, V_i)$, where $k_1$ is the starting keyframe. The beam search explores possible keyframe paths $\mathcal{P}_i = \{k_1, k_2, \dots, \hat{k}_n\}$ satisfying:
    \[
    V(k_j) = V_i, \quad t_{k_1} < t_{k_2} < \dots < t_{\hat{k}_n},
    \]
    subject to a maximum temporal gap $\tau$ between consecutive events.
    
    At each beam search step, we retain the top-$b$ partial paths based on cumulative similarity scores. The search objective is to maximize the total event alignment score:
    \[
    EventScore(S_i) = \max_{\mathcal{P}_i} \sum_{j=1}^{n} SimScore_j(k_j),
    \]
    where $\mathcal{P}_i$ is a valid temporally ordered keyframe path in video $V_i$.
    \item \textbf{Reranking Segments:}
    The final relevance score combines the semantic (context) and visual (event) components through a weighted sum:
    \[
    FinalScore(S_i)=\alpha \cdot EventScore(S_i) + (1-\alpha) \cdot ContextScore(S_i)
    \]
    where $\alpha \in [0;1]$ controls the balance between visual similarity and contextual alignment. The top-ranked segments according to $FinalScore(S_i)$ are returned as the final retrieval results.
    
\end{enumerate}

\subsection{Google Image Search}
To address "out-of-distribution" (OOD) queries that our embedding model cannot accurately interpret, we have integrated a context enrichment workflow using Google Search.
Specifically, when this feature is enabled, the user inputs their text query into a dedicated search bar. The system immediately uses Google Search to retrieve the top $k$ most relevant images from the Internet. These images are presented to the user, who then selects the single image that most accurately reflects their intent.
Once selected, this image is processed by the CLIP (Contrastive Language-Image Pre-Training) model to generate an embedding vector. This visually semantic-rich vector is then used to supplement and augment the original query context. This validated combination of text and image allows the system to retrieve the semantically relevant video far more accurately, even for complex OOD queries.

While this approach significantly enhances semantic robustness, it introduces a key limitation: variable latency. The process is inherently slower than standard text retrieval due to its reliance on two non-trivial overheads: the external API call to Google Search and the requisite step of manual user selection. This design represents a deliberate trade-off, prioritizing access to external knowledge for difficult OOD queries over the immediate speed of internal-only retrieval. Consequently, the feature is implemented as a user-initiated fallback mechanism rather than a default search pathway.
\section{System usage}
\subsection{Overall UI}
\begin{figure}[h]
    \centering
    \includegraphics[width=\textwidth]{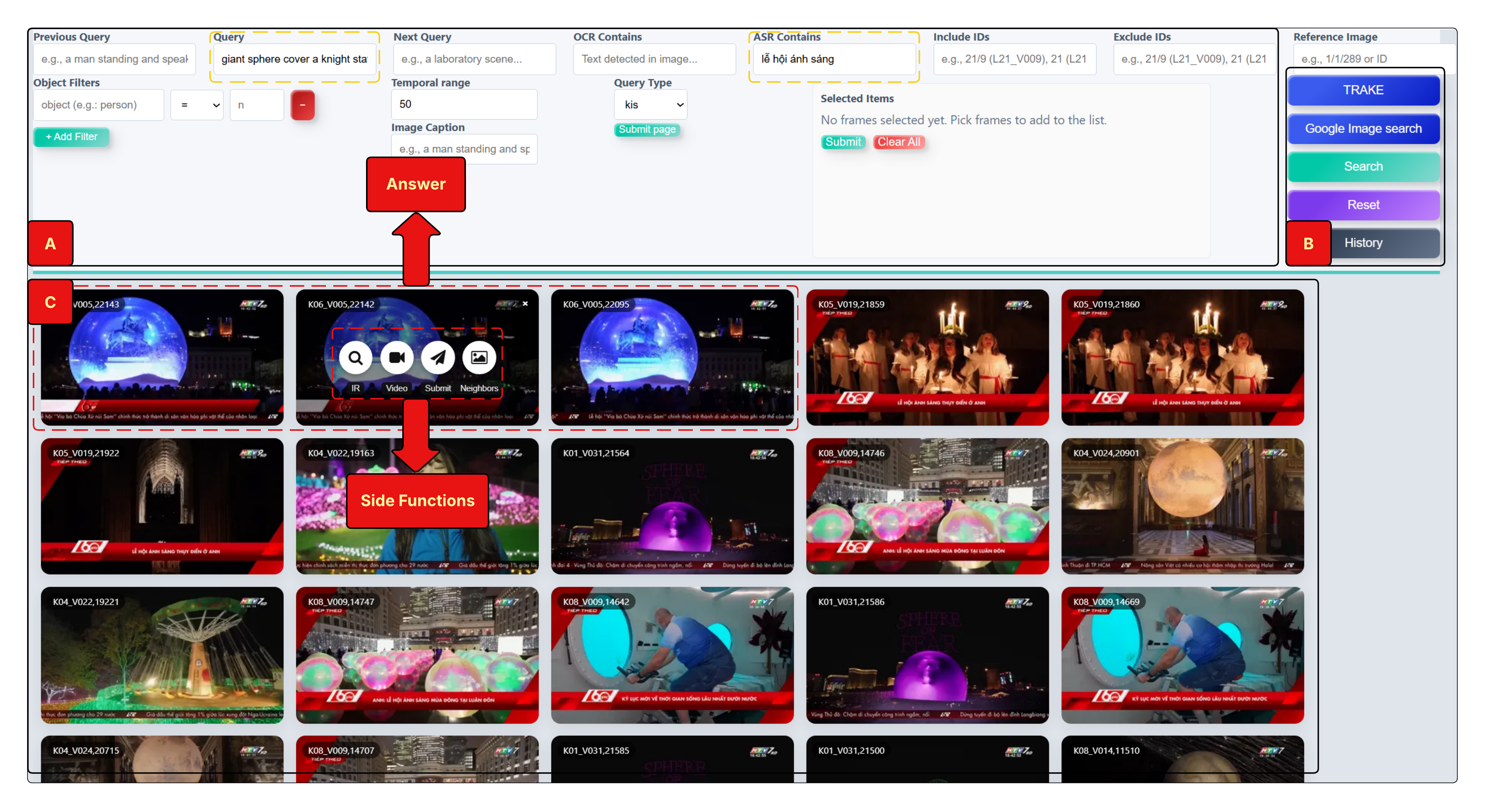}
    \caption{Interface of the video frame search and extraction system. 
    Section (A) contains query input and filters, 
    section (B) provides control buttons for actions such as Google Image Search, TRAKE, and Reset,
    while section (C) displays retrieved video frames in a grid layout.}
    \label{fig:video_frame_search_ui}
\end{figure}

To support intuitive interaction with the multimodal video database, we develop an interface for both exploratory and targeted retrieval (Figure~\ref{fig:video_frame_search_ui}). The UI comprises three components.

\textbf{Query Panel (A).} Users compose hybrid queries by combining metadata filters with keyword or semantic text search across objects, OCR, ASR transcripts, and captions. Temporal constraints (sequential or range-based) enable event-level retrieval. Options such as \textit{Include/Exclude IDs}, \textit{Query Type}, and \textit{Query Pack} support iterative refinement and QA-style tasks. A \textit{Selection Box} maintains chosen keyframes, with \textit{Submit} and \textit{Clear All} for rapid iteration.

\textbf{Query Execution Module (B).} This module manages query operations: \textit{Search} performs hybrid retrieval, \textit{Google Image Search} triggers contextual query expansion (Section~\ref{gis}), and \textit{TRAKE} activates the multi-event temporal retrieval pipeline (Section~\ref{TRAKE_UI}).

\textbf{Results Display Area (C).} Retrieved keyframes appear in a labeled grid (e.g., K21/V0001/0155). Selecting a keyframe opens a detailed viewer (Figure~\ref{fig:video_frame_search_ui}) with contextual navigation. The viewer shows the main frame with a temporal filmstrip for assessing continuity, and provides tools for relevant image search, frame exclusion, and adding frames to the \textit{Selected Items} for submission.

\subsection{Trake Search}
\label{TRAKE_UI}
\begin{figure}[t]
    \centering
    \includegraphics[width=\linewidth]{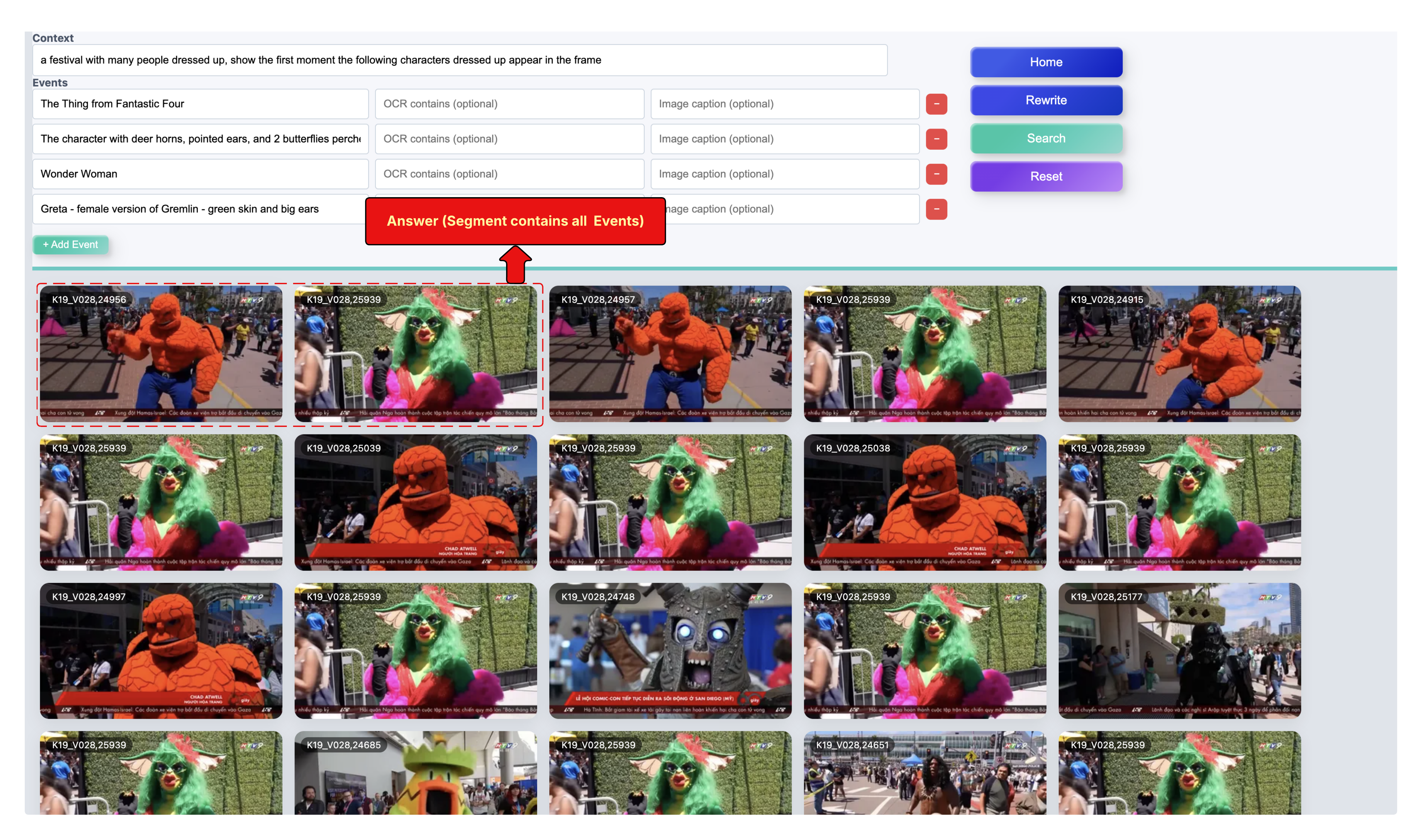}
    \caption{TRAKE User Interface for multi-event video retrieval. Users define a high-level context and an ordered sequence of events, each with optional metadata filters. The system retrieves video segments satisfying the specified temporal relationships, highlighting relevant matches (red dashed boxes).}
    \label{fig:Trake UI}
\end{figure}

Figure~\ref{fig:Trake UI} illustrates the TRAKE Search Interface, designed to facilitate multi-event temporal retrieval. The interface allows users to define a context and a sequence of event queries, each optionally constrained by OCR or caption filters. Once submitted, the system executes temporal reasoning across candidate videos to identify segments that align with the specified event order. 

\subsection{Google Image Search}

\begin{figure}[h]
    \centering
    \includegraphics[width=\textwidth]{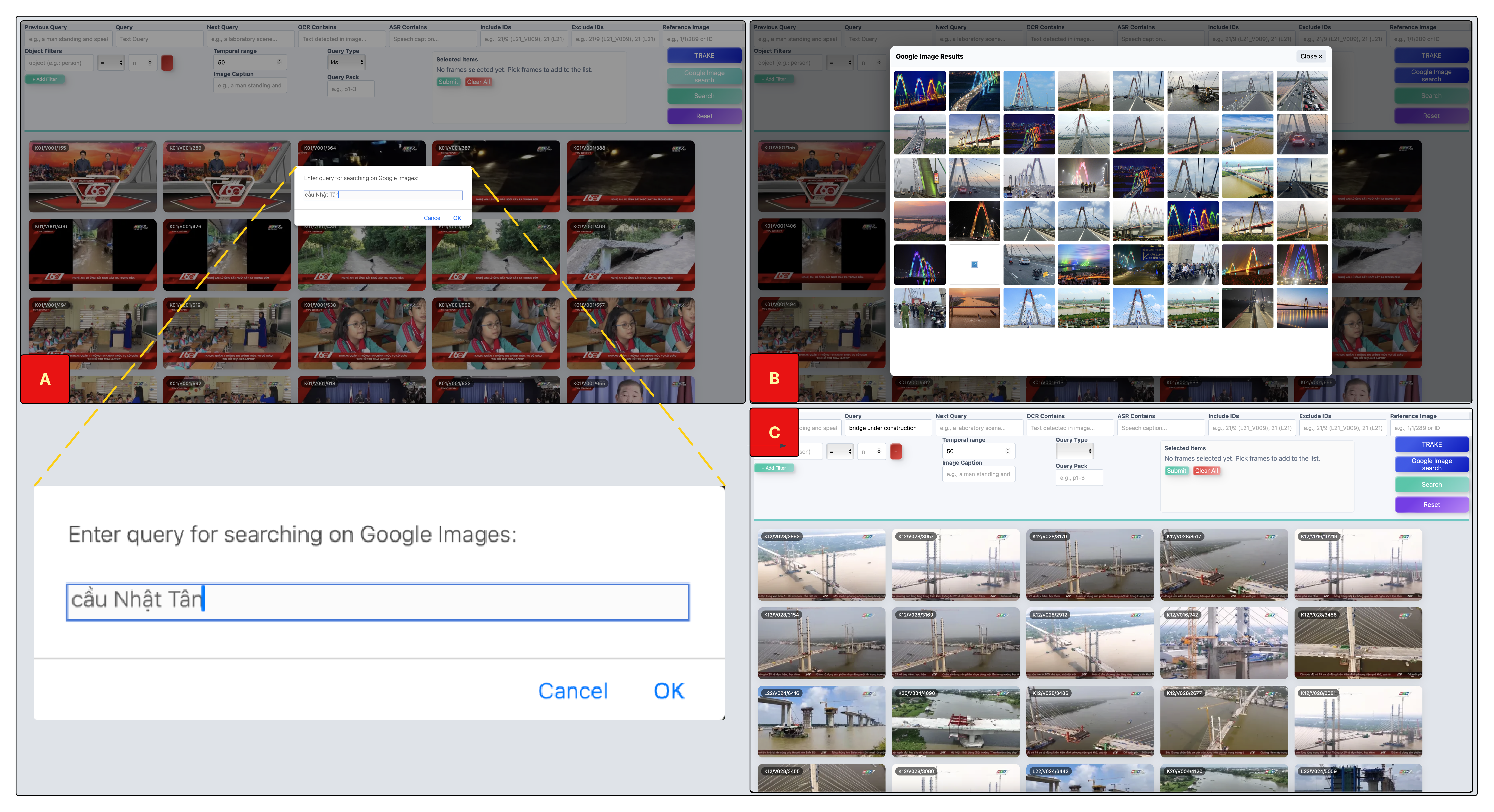}
    \caption{The Visual Query Enrichment workflow integrating Google Image Search. (A) The user initiates a text-based query (e.g., "Nhat Tan bridge") via a dialog box. (B) The system utilizes an API to retrieve image results, from which the user selects a relevant visual reference. (C) The selected image is then integrated into the system to refine and enrich the context for the final video search query.}
    \label{fig:gg_res}
\end{figure}
\label{gis}

The Google Image Search functionality is illustrated in Figure~\ref{fig:gg_res}. As shown, when a user utilizes this feature, the system displays a search bar where the user can input the desired object (for example, the Nhat Tan Bridge here). The system then uses the Google API to return corresponding images of that object. The user subsequently selects the most suitable image. Following this, the system integrates the information from the chosen image to enrich the context for the query.

\section{Performance and Results}
The effectiveness of our \textbf{MADTempo} system was validated during its participation in the Ho Chi Minh City AI Challenge 2025~\cite{AIChallenge2025}. Our team, \textbf{AIO\_Trình}, successfully advanced from the Preliminary Round with an official score of \textbf{75.4}.

In the subsequent Final Round, the system was evaluated across four distinct retrieval task categories, ultimately securing a \textbf{Very good} overall performance ranking. The detailed performance breakdown for the Final Round is as follows:
\begin{itemize}
    \item \textbf{TKIS tasks (Text-to-Keyframe):} Good
    \item \textbf{VKIS tasks (Visual-to-Keyframe):} Very good
    \item \textbf{TRAKE tasks (Temporal Retrieval):} Excellent
    \item \textbf{QA tasks (Query Answering):} Good
\end{itemize}

We attribute the \textbf{Excellent} performance in the TRAKE tasks directly to our robust multi-event temporal search algorithm (detailed in Section~\ref{sec:trake}), which confirms its capability to handle complex event sequences. Furthermore, the \textbf{Very good} result in VKIS tasks highlights the successful integration of our Google Image Search–based fallback mechanism (Section~\ref{gis}) for handling challenging out-of-distribution (OOD) visual queries.

\section{Conclusion}
This paper presented \textbf{MADTempo}, an interactive video retrieval system centered on an efficient temporal search algorithm for multi-event queries. Our method enables consistent retrieval of complex action sequences that are difficult for conventional approaches.

Although our prototype employs GPT-5 for query enhancement, the module is fully \emph{LLM-agnostic}. Any open-source model can be substituted without altering the architecture, ensuring controllable deployment costs, long-term stability, and broad accessibility. Because this component functions solely as a lightweight query refiner, it does not affect the core temporal retrieval pipeline.

We further incorporate Google Image Search as an optional user-triggered enrichment tool, trading minor latency for valuable external visual grounding—particularly effective for out-of-distribution queries—while leaving standard retrieval performance unaffected.

Early user evaluations show that MADTempo’s interface and temporal reasoning significantly accelerate search efficiency. Future work will explore richer multimodal querying, such as voice-based input and a sketch-enabled canvas for structured scene retrieval.
%
%
%

\bibliographystyle{splncs04}
\bibliography{mybibliography}

\end{document}